\documentclass[runningheads]{llncs}
\usepackage{graphicx}
\usepackage{comment}
\usepackage{amsmath,amssymb} % define this before the line numbering.
\usepackage{color}
\usepackage{url}

\usepackage{amsfonts}
\usepackage{siunitx}

\usepackage[ruled,vlined]{algorithm2e}

\usepackage{booktabs}
\usepackage{epsfig}
\usepackage{etoolbox}
\usepackage{float}

\usepackage{array}
\newcolumntype{C}[1]{>{\centering\arraybackslash}p{#1}}

\usepackage{caption}
\usepackage{subcaption}
\def\E{{\rm E}}
\def\d{\frac{\partial}{\partial \theta}}
\def\N{\mathcal{N}}

\newcommand*\samethanks[1][\value{footnote}]{\footnotemark[#1]}

\usepackage{xcolor}

\begin{document}
\pagestyle{headings}
\mainmatter

\title{Learning Multi-layer Latent Variable Model via Variational Optimization of Short Run MCMC for Approximate Inference} % Replace with your title

\titlerunning{Variational Optimization of Short Run MCMC for Approximate Inference} 
\authorrunning{E. Nijkamp, B. Pang, T. Han, L. Zhou, S.-C. Zhu, and Y. N. Wu}
\author{Erik Nijkamp\inst{1}\thanks{Equal contribution.}\and
	Bo Pang\inst{1}\samethanks\and
	Tian Han\inst{2}\and
	Linqi Zhou\inst{1}\and\\
	Song-Chun Zhu\inst{1}\and
	Ying Nian Wu\inst{1}
}
\institute{University of California, Los Angeles\\
	\email{\{enijkamp,bopang,linqi.zhou\}@ucla.edu},
	\email{\{sczhu,ywu\}@stat.ucla.edu}\and
	Stevens Institute of Technology\\
	\email{than6@stevens.edu}
}
%******************
\maketitle

\begin{abstract}
This paper studies the fundamental problem of learning deep generative models that consist of multiple layers of latent variables organized in top-down architectures. Such models have high expressivity and allow for learning hierarchical representations. Learning such a generative model requires inferring the latent variables for each training example based on the posterior distribution of these latent variables. The inference typically requires Markov chain Monte Caro (MCMC) that can be time consuming.  In this paper, we propose to use noise initialized non-persistent short run MCMC, such as finite step Langevin dynamics initialized from the prior distribution of the latent variables, as an approximate inference engine, where the step size of the Langevin dynamics is variationally optimized by minimizing the Kullback-Leibler divergence between the distribution produced by the short run MCMC  and the posterior distribution. Our experiments show that the proposed method outperforms variational auto-encoder (VAE) in terms of reconstruction error and synthesis quality. The advantage of the proposed method is that it is simple and automatic without the need to design an inference model.
\end{abstract}

%%%%%%%%% BODY TEXT
\section{Introduction}

Deep generative models have seen many applications such as image and video synthesis, and unsupervised or semi-supervised learning. Such models usually consist of one or more layers of latent variables organized in top-down architectures. Learning such latent variable models from training examples is a fundamental problem, and this paper studies this problem for top-down models with multiple layers of latent variables. Such models have high expressivity and allow for learning hierarchical representations.

Learning latent variable models requires inferring the latent variables based on their joint posterior distribution, i.e., the conditional distribution of the latent variables given each observed example. The inference typically requires Markov chain Monte Carlo (MCMC) such as Langevin dynamics~\cite{langevin1908theory} or Hamiltonian Monte Carlo (HMC)~\cite{neal2011mcmc}. Such MCMC posterior sampling can be time consuming and difficult to scale up. The convergence of MCMC sampling in finite time is also questionable, especially if the posterior distribution is multi-modal. 

An alternative to MCMC posterior sampling is variational inference, such as variational auto-encoder (VAE)~\cite{kingma2013auto,rezende2014stochastic}, which learns an extra inference network that maps each input example to the approximate posterior distribution.  
Despite the success of VAE, it has the following shortcomings. (1) It requires a separate inference model with a separate set of parameters. These parameters are to be learned together with the parameters of the generative model. (2) The design of the inference model is not automatic, especially for generative models with multiple layers of latent variables, which may have complex relationships governed by their joint posterior distribution. It is a highly non-trivial task to design an inference model to adequately capture the explaining-away competitions and bottom-up and top-down interactions between layers of latent variables~\cite{maaloe2019biva,sonderby2016ladder}. 

The goal of this paper is to completely do away with a separate inference model. Specifically, we propose to use noise initialized non-persistent short run MCMC~\cite{nijkamp2019learning}, such as finite step Langevin dynamics, as an approximate inference engine.  In the learning process, for each training example, we always initialize such a short run MCMC from the prior distribution of the latent variables, such as Gaussian or uniform noise distribution, and run a fixed finite number (e.g., 25) of steps.  Thus the short run MCMC is non-persistent. 
In agreement with the philosophy of variational inference, we accept the approximate nature of short run MCMC, and we optimize the step size, or in general, algorithmic hyper-parameters of the short run MCMC, by minimizing the Kullback-Leibler divergence between the approximate distribution produced by the short run MCMC and the  posterior distribution. This is a variational optimization, except that the variational parameter is the step size.  Our experiments show that the proposed method outperforms VAE for multi-layer latent variable models in terms of reconstruction error and synthesis quality. 

One major advantage of the proposed method is that it is simple and automatic. For models with multiple layers of latent variables that may be organized in complex top-down architectures, the gradient computation in Langevin dynamics is automatic on modern deep learning platforms. Such dynamics naturally integrates explaining-away competitions and bottom-up and top-down interactions between multiple layers of latent variables. It thus enables researchers to explore flexible generative models without dealing with the challenging task of designing and learning the inference models.  

One class of generative models that are of particular interest are biologically plausible models, such as Boltzmann machine~\cite{hinton1985boltzmann} and the generation model of the Helmholtz machine~\cite{hinton1995wake}, where each node is a latent variable. With such a large number of latent variables, designing an inference network to regulate the bottom-up and top-down flows of information as well as lateral inhibitions becomes a daunting task. However, short run MCMC is automatic, natural, and biologically plausible as it may be related to attractor dynamics~\cite{hopfield1982neural,amit1989world,poucet2005attractors}. 

\section{Contributions and related work} 

The following are contributions of our paper. 
\begin{itemize}
	\item We propose short run MCMC for approximate inference of latent variables in deep generative models. 
	\item We provide a method to determine the optimal step size, or in general, hyper-parameters of the short run MCMC.
	\item We demonstrate learning of multi-layer latent variable models with high quality samples and reconstructions.
\end{itemize}

The following are themes related to our work. 

(1) {\em Variational inference.} As mentioned above, VAE~\cite{kingma2013auto,rezende2014stochastic,sonderby2016ladder,gregor2015draw} is the prominent method for learning generator network. Our short run MCMC can be considered an inference model, except that it is intrinsic to the generative model in that it is based on the parameters of the generative model. Thus there is little mismatch between the inference process and the generative model, even at the beginning stage of the learning algorithm. Only algorithmic hyper-parameters such as step size are optimized by variational criterion. It is particularly convenient for models with multiple layers of latent variables, whereas designing variational inference models for such generative models can be a highly non-trivial task. 

(2) {\em Alternating back-propagation.} \cite{han2017abp} proposes to learn the generator network by maximum likelihood, and the learning algorithm iterates the following two steps: (a) inferring the latent variables by Langevin dynamics that samples from the posterior distribution of the latent variables. (b) updating the model parameters based on the inferred latent variables. Both steps involve gradient computations based on back-propagation. Similar training scheme has been developed and extended to model flexible latent prior as in \cite{qiu2019almond,pang2020learning} and spatial-temporal data as in \cite{han2019learning,xie2019abptt}. \cite{chen18ctf} also leverages Langevin dynamics for posterior sampling which is however initialized from samples produced by an inference network. In the training stage, in step (a), the Langevin dynamics is initialized from the samples produced in the previous learning epoch. This is usually called persistent chain in the literature~\cite{pcd}. In our work, in step (a), we always initialize the finite-step (e.g., 25-step) Langevin updates from the prior noise distribution. This can be called non-persistent chain. The following are advantages of our method based on non-persistent short run MCMC as compared to methods based on persistent chain. (1) The short run MCMC can be viewed as an inference model whose hyper-parameters can be optimized based on variational criterion. This strikes a middle ground between MCMC and variational inference. (2) Theoretical underpinning of the learning method based on short run MCMC is much cleaner. (3) In both training and testing stages, the same short run MCMC is used. 

(3) {\em Short run MCMC for energy-based model.} Recently \cite{nijkamp2019learning} proposes to learn short run MCMC for energy-based model (EBM). An EBM is in the form of an unnormalized probability density function, where the log-density or the energy function is parametrized by a bottom-up neural network. \cite{nijkamp2019learning} shows that it is possible to learn noise initialized non-persistent short run MCMC such as 100-step Langevin dynamics that can generate images of high synthesis quality.  Our method follows a similar strategy, but it is intended for approximately sampling from the posterior distribution of latent variables. % Our optimization method can help reduce the gap between the short run MCMC and the corresponding EBM, so that the EBM may also be properly learned. 

(4) {\em Attractor dynamics.} In computational neuroscience, the dynamics of the neuron activities is often modeled by attractor dynamics \cite{hopfield1982neural,amit1989world,poucet2005attractors}. However, the objective function of the attractor dynamics is often implicit, thus it is unclear what is the computational problem that the attractor dynamics is solving. For the attractor dynamics to be implemented in real time, the dynamics is necessarily a short run dynamics. Our short run MCMC is guided by a top-down model with a well-defined posterior distribution of the latent variables. It may be connected to the attractor dynamics and help us understand the latter. We shall explore this direction in future work.

\section{Top-down model with multi-layer latent variables}

\subsection{Joint, marginal, and posterior distributions}

Let $x$ be the observed example, such as an image. Let $z$ be the latent variables, which may consist of latent variables at multiple layers organized in a top-down architecture. 

The joint distribution of $(x, z)$ is $p_\theta(x, z)$, where $\theta$ consists of model parameters. The marginal distribution of $x$ is 
$
p_\theta(x) = \int p_\theta(x, z) dz. 
$
Given $x$, the inference of $z$ can be based on the posterior distribution $p_\theta(z|x) = p_\theta(x, z)/p_\theta(x)$. 

The generator network assumes a $d$-dimensional noise vector $z$ at the top-layer. The prior distribution $p(z)$ is known, such as $z \sim \N(0, I_d)$, where $I_d$ is the $d$-dimensional identity matrix. Given $z$, $x = g_\theta(z) + \epsilon$, where $g_\theta(z)$ is a top-down convolutional neural network (sometimes called deconvolutional network due to the top-down nature), where $\theta$ consists of all the weight and bias terms of this top-down network. $\epsilon$ is usually assumed to be Gaussian white noise with mean 0 and variance $\sigma^2$. Thus $p_\theta(x|z)$ is such that $[x|z] \sim \N(g_\theta(z), \sigma^2 I_D)$, where $D$ is the dimensionality of $x$. For this model 
\begin{align}
\log  p_\theta(x, z) &= \log [p(z) p_\theta(x|z) ] \\
&= -\frac{1}{2} \left[ \|z\|^2 + \|x - g_\theta(z)\|^2/\sigma^2\right] + c, \label{eq:L}
\end{align}
where $c$ is a constant independent of $\theta$. 

 In this paper, we are mainly concerned with multi-layer generator network. While it is computationally convenient to have a single latent noise vector at the top layer, it does not account for the fact that patterns can appear at multiple layers of compositions or abstractions (e.g., face $\rightarrow$ (eyes, nose, mouth) $\rightarrow$ (edges, corners) $\rightarrow$ pixels), where  variations and randomness occur at multiple layers. To capture such a hierarchical structure, it is desirable to introduce multiple layers of latent variables organized in a top-down architecture. Specifically, we have $z = (z_l, l = 1, ..., L)$, where layer $L$ is the top layer, and layer $1$ is the bottom layer above $x$. For notational simplicity, we let $x = z_0$. We can then specify $p_\theta(z)$ as 
\begin{align}
p_{\theta}(z) = p_{\theta}(z_L) \prod_{l=0}^{L-1}p_{\theta}(z_l | z_{l+1}).
\end{align}
One concrete example is 
$z_L \sim \N(0, I)$, 
$[z_l|z_{l+1}] \sim \N(\mu_l(z_{l+1}), \sigma_l^2(z_{l+1})),  \; l = 0, ..., L-1$.
where $\mu_l()$ and $\sigma_l^2()$ are the mean vector and the diagonal variance-covariance matrix of $z_l$ respectively, and they are functions of $z_{l+1}$. $\theta$ collects all the parameters in these functions. $p_\theta(x, z)$ can be obtained similarly as in Equation (\ref{eq:L}). 

\subsection{Learning and inference} 

Let $p_{\rm data}(x)$ be the data distribution that generates the example $x$. The learning of parameters $\theta$ of $p_\theta(x)$ can be based on 
$
\min_\theta {\rm KL}(p_{\rm data}(x)\| p_\theta(x)),
$
where ${\rm KL}(p\|q) = \E_p[\log (p(x)/q(x))]$ is the Kullback-Leibler divergence between $p$ and $q$ (or from $p$ to $q$ since ${\rm KL}(p\|q)$ is asymmetric). If we observe training examples $\{x_i, i = 1, ..., n\} \sim p_{\rm data}(x)$, the above minimization can be approximated by maximizing the log-likelihood 
\begin{align}
L(\theta)  = \frac{1}{n} \sum_{i=1}^{n} \log p_\theta(x_i), 
\end{align}
which leads to the maximum likelihood estimate (MLE). 

The gradient of the log-likelihood, $L'(\theta)$, can be computed according to the following identity: 
\begin{align} 
\d \log p_\theta(x) &= \frac{1}{p_\theta(x)} \d p_\theta(x) \\
 % &= \frac{1}{p_\theta(x)} \int \d p_\theta(x, z) dz \\
&= \int \d \log p_\theta(x, z) \frac{p_\theta(x, z)}{p_\theta(x)} dz\\
&= \E_{p_\theta(z|x)} \left[ \d \log p_\theta(x, z)\right]. 
\end{align}
The above expectation can be approximated by Monte Carlo samples from $p_\theta(z|x)$. The MLE learning can be accomplished by gradient descent. Each learning iteration updates $\theta$ by 
\begin{align} 
\theta_{t+1} = \theta_t + \eta_t \frac{1}{n} \sum_{i=1}^{n} \E_{p_{\theta_t}(z_i|x_i)}\left[ \d \log p_{\theta}(x_i, z_i) \mid_{\theta = \theta_t} \right],  \label{eq:T0}
\end{align}
where $\eta_t$ is the step size or learning rate, and $\E_{p_{\theta_t}(z_i|x_i)}$  can be approximated by Monte Carlo sampling from $p_{\theta_t}(z_i|x_i)$.

\section{Short run MCMC for approximate inference}

\subsection{Langevin dynamics} 

Sampling from $p_\theta(z|x)$ usually requires MCMC. One convenient MCMC is Langevin dynamics~\cite{langevin1908theory}, which iterates 
\begin{align} 
z_{k+1} = z_k + s \frac{\partial}{\partial z} \log p_\theta(z_k|x) + \sqrt{2s} \epsilon_k, 
\end{align}
where $\epsilon_k \sim \N(0, I)$,  $k$ indexes the time step of the Langevin dynamics, and $s$ is the step size. The Langevin dynamics consists of a gradient descent term on $-\log p(z|x)$. In the case of generator network, it amounts to gradient descent on $\|z\|^2 /2+ \|x - g_\theta(z)\|^2/2\sigma^2$, which is penalized reconstruction error. The Langevin dynamics also consists of a white noise diffusion term $\sqrt{2s} \epsilon_k$ to create randomness for sampling from $p_\theta(z|x)$. 

For small step size $s$, the marginal distribution of $z_k$ will converge to $p_\theta(z|x)$ as $k \rightarrow \infty$ regardless of the initial distribution of $z_0$. More specifically, let $q_k(z)$ be the marginal distribution of $z_k$ of the Langevin dynamics, then ${\rm KL}(q_k(z) \| p_\theta(z|x))$ decreases monotonically to 0, that is, by increasing $k$, we reduce ${\rm KL}(q_k(z) \| p_\theta(z|x))$ monotonically~\cite{cover2012elements}.  %Thus MCMC is consistent with variational inference. Both seek to minimize ${\rm KL}(q(z) \| p_\theta(z|x))$ over $q$, which is within a certain class. 

\subsection{Noise initialized short run MCMC} 

It is impractical to run long chains to sample from $p_\theta(z|x)$. We thus propose the following short run MCMC as inference dynamics, with a fixed small $K$ (e.g., $K = 25$), 
\begin{align} 
z_0 \sim p(z),
 z_{k+1} = z_k + s \frac{\partial}{\partial z} \log p_\theta(z_k|x)  + \sqrt{2s} \epsilon_k, \; k = 1, ..., K,  \label{eq:S}
\end{align}
where $p(z)$ is the prior noise distribution of $z$. 

We can write the above short run MCMC as 
\begin{align}
z_0 \sim p(z), \; z_{k+1} = z_k + s R(z_k) + \sqrt{2s} \epsilon_k, \; k = 1, ..., K, 
\end{align}
 $R(z) = \frac{\partial}{\partial z} \log p_\theta(z|x)$, where we omit $x$ and $\theta$ in $R(z)$ for simplicity of notation. For finite $K$, this dynamics is a $K$-layer noise-injected residual network~\cite{he2016resnet}, or $K$-step noise-injected RNN~\cite{rumelhart1986rnn,hochreiter1997lstm}. It may also be compared to flow-based inference model~\cite{dinh2014nice,germain2015made,dinh2016density,kingma2016iaf,kingma2018glow}, except we do not learn a separate inference model. 
 
To further simplify the notation, we may write the short run MCMC as 
\begin{align}
z_0 \sim p(z), \; z_K = F(z_0, \epsilon),
\end{align}
where $\epsilon = (\epsilon_k, k = 1, ..., K)$, 
and $F$ composes the $K$ steps of Langevin updates. 

Let the distribution of $z_K$ be $q_s(z)$, where we include the notation $s$ to make it explicit that the distribution of $z_K$ depends on the step size $s$. Recall that the distribution of $z_K$ also depends on $x$ and $\theta$, so that in full notation, we may write $q_s(z)$ as $q_{s, \theta}(z|x)$. 

For short run MCMC (\ref{eq:S}), the gradient term usually dominates the noise term, and most of the randomness comes from $z_0 \sim p(z)$.  Given $\epsilon$, $z_K$ is a deterministic transformation of $z_0$. Assuming this transformation is invertible, and let $z_0 = F^{-1}(z_k, \epsilon)$.  Let $q_s(z|\epsilon)$ be the conditional distribution of $z_K$ given $\epsilon$. By change of variable, 
\begin{align} 
 q_s(z|\epsilon) = p(F^{-1}(z, \epsilon))|{\rm det}(d F^{-1}(z, \epsilon)/d z)|. 
\end{align}
Then 
\begin{align}
q_s(z) = \int q_s(z|\epsilon) p(\epsilon) d \epsilon = \E_{p(\epsilon)}[q_s(z|\epsilon)], 
\end{align}
which can be approximated by Monte Carlo sampling from $p(\epsilon)$, i.e., the iid $\N(0, I)$ distribution. 

For our method, we never need to compute $F^{-1}$, because we only need to compute $\E[h(z_K)] = \E_{q_s(z)}[h(z)]$ for a given function $h$, and 
\begin{align} 
\E_{q_s(z)}[h(z)] = \E_{p(z_0) p(\epsilon)} [h(F(z_0, \epsilon))]. \label{eq:C1}
\end{align}
In particular, we need to compute the entropy of $q_s(z)$ for variational optimization of step size $s$. The entropy is the negative of 
\begin{align}
    \E_{q_s(z)} [ \log q_s(z)] & =  \E_{p(z_0)p(\epsilon)} [ \log \E_{p(\epsilon)} (q_s(F(z_0, \epsilon)|\epsilon))]\\
   &=  \E_{p(z_0)p(\epsilon)}[ \log \E_{p(\epsilon)}(p(z_0)/|\det(d F(z_0, \epsilon)/dz_0)|)], \label{eq:C2}
\end{align}
where the expectations can be approximated by Monte Carlo sampling from the known prior distribution of $z_0$ and the known noise distribution of $\epsilon$. In the above computation, we need to compute the determinant of the Jacobian $d F(z_0, \epsilon)/d z_0$. Fortunately, on modern deep learning platforms, such computation is easily feasible even if the dimension of $z_0$ is very high. Specifically, after computing the matrix $d F(z_0, \epsilon)/d z_0$, we can compute the eigenvalues of $d F(z_0, \epsilon)/d z_0$, so that the log-determinant is the sum of the log of the eigenvalues. 

As to the invertibility of $F$, in our experience, the eigenvalues of $d F(z_0, \epsilon)/d z_0$ are always away from 0, suggesting that $z_K = F(z_0, \epsilon)$ is locally invertible. Moreover,  different $z_0$ always lead to different $z_K = F(z_0, \epsilon)$,   suggesting that $F$ is globally invertible. Again, our method does not require inverting $F$. 

\subsection{Variational optimization of step size}\label{sec:ss}

We want to optimize the step size $s$ so that $q_s(z)$ best approximates the posterior $p_\theta(z|x)$. This can be accomplished by 
\begin{align}
\min_s {\rm KL}(q_s(z) \| p_\theta(z|x)). 
\end{align} 
This is similar to variational approximation, with step size~$s$ being the variational parameter. 
\begin{align}
{\rm KL}(q_s(z) \| p_\theta(z|x))& = \E_{q_s(z)}[\log q_s(z) - \log p_\theta(x, z)]  + \log p_\theta(x), 
\end{align}
where the last term $\log p_\theta(x)$ is independent of $s$. The computation of the first two terms is explained in the previous subsection. See equations (\ref{eq:C1}) and (\ref{eq:C2}).  

While we can optimize the step size $s$ for each example $x$, in our work, we optimize over an overall $s$ that is shared by all the examples. Reverting to the full notation $q_{s, \theta}(z|x)$ for $q_s(z)$, this means we minimizes 
\begin{align}
\frac{1}{n}\sum_{i=1}^{n}  {\rm KL}(q_{s, \theta}(z_i|x_i) \| p_{\theta}(z_i|x_i)) \label{eq:M}
\end{align} 
over $s$. The minimization can be accomplished by a grid search, or by gradient descent (the gradient is still computable on modern deep learning platforms). 

Instead of using a constant step size $s$ for all $k$, we may also optimize over varying step sizes $s_k, k = 1, ..., K$. We leave it to future work.

The main computational burden in optimizing algorithmic hyper-parameters such as step size comes from the computation of the entropy of $q_{s, \theta}(z_i|x_i)$. In this paper, we compute it rigorously to make the learning principled. In future work, we shall explore efficient approximate methods to optimize short run MCMC. 

%We may also employ approximations or surrogates of this entropy term for more efficient computation. In fact, we suspect that this entropy term may not be very crucial, because the short run MCMC is initialized from a diffused noise distribution, and there is a noise or diffusion term in each step of the Langevin dynamics. Thus the randomness in $q_{s, \theta}(z_i|x_i)$ may not be affected much by algorithmic hyper-parameters. 

\subsection{Learning with short run MCMC} 

A learning iteration consists of the following two steps. (1) Update $s$ by minimizing (\ref{eq:M}). (2) Update $\theta$ by 
\begin{align} 
\theta_{t+1} = \theta_t + \eta_t \frac{1}{n} \sum_{i=1}^{n} \E_{q_{s, \theta_t}(z_i|x_i)}\left[ \d \log p_{\theta}(x_i, z_i) \mid_{\theta = \theta_t} \right],  \label{eq:T}
\end{align}
where $\eta_t$ is the learning rate, $\E_{q_{s, \theta_t}(z_i|x_i)}$ (here we use the full notation $q_{s, \theta}(z|x)$ instead of the abbreviated notation $q_s(z)$) can be approximated by sampling from $q_{s, \theta_t}(z_i|x_i)$ using the noise initialized $K$-step Langevin dynamics. Compared to MLE learning algorithm (\ref{eq:T0}), we replace $p_{\theta_t}(z|x)$ by $q_{s, \theta}(z|x)$, and fair Monte Carlo samples from $q_{s, \theta}(z|x)$ can be obtained by short run MCMC.

The learning procedure is summarized in Algorithm~\ref{algo:short}. Note, we only optimize $s$ every $T_s$ iterations, so that it does not incur much computational burden.\\
\\
\begin{algorithm}[H]
	\SetKwInOut{Input}{input} \SetKwInOut{Output}{output}
	\DontPrintSemicolon
	\Input{Training examples~$\{x_i \}_{i=1}^n$, learning iterations~$T$, step size updating interval $T_s$, learning rate~$\eta$, initial parameters~$\theta_0$,  batch size~$m$, number of steps $K$, initial step size $s$.}
	\Output{Parameters $\theta_{T}$.}
	\For{$t = 0:T-1$}{			
		\smallskip
		1. Draw observed examples $\{ x_i \}_{i=1}^m$. \;
		2. Draw latent vectors $\{ z_{i,0} \sim p(z) \}_{i=1}^m$.\;
		3. Infer $\{ z_{i,K} \}_{i=1}^m$ by $K$ steps of dynamics (\ref{eq:S}) with step size $s$.\;
		4. Update $\theta$ according to (\ref{eq:T}).\;
		5. Every $T_s$ iterations, update $s$ by minimizing (\ref{eq:M}).\;
	}
	\caption{Learning with short run MCMC.}
	\label{algo:short}
\end{algorithm}

\subsection{Theoretical underpinning} 

Given $\theta_t$, the updating equation (\ref{eq:T}) is a one step gradient ascent on
\begin{align} 
Q_s(\theta) =  \frac{1}{n} \sum_{i=1}^{n} \E_{q_{s, \theta_t}(z_i|x_i)}\left[ \log p_{\theta}(x_i, z_i) \right].
\end{align}

Compared to the log-likelihood  in MLE learning, ${L(\theta) = \frac{1}{n} \sum_{i=1}^{n} \log p_\theta(x)}$, 
\begin{align} 
Q_s(\theta) &= L(\theta) + \frac{1}{n} \sum_{i=1}^{n}  \E_{q_{s, \theta_t}(z_i|x_i)}\left[ \log p_{\theta}(z_i|x_i) \right]\\
&= L(\theta) - \frac{1}{n}\sum_{i=1}^{n}  {\rm KL}(q_{s, \theta_t}(z_i|x_i) \| p_{\theta}(z_i|x_i)) \\
& +\frac{1}{n}\sum_{i=1}^{n}   \E_{q_{s, \theta_t}(z_i|x_i)}[\log q_{s, \theta_t}(z_i|x_i)]. 
\end{align}
Since the last term has nothing to do with $\theta$, gradient ascent on $Q_s(\theta)$ is equivalent to gradient ascent of $\tilde{Q}_s(\theta) = L(\theta) - \frac{1}{n}\sum_{i=1}^{n}  {\rm KL}(q_{s, \theta_t}(z_i|x_i) \| p_{\theta}(z_i|x_i))$, which is a lower bound of $L(\theta)$.  $\tilde{Q}_s(\theta)$ is a perturbation of $L(\theta)$. At $\theta_t$, the optimization over $s$ by minimizing (\ref{eq:M}) is to minimize this perturbation. 

Thus a learning iteration can be interpreted as a joint maximization of $\tilde{Q}_s(\theta)$ over $s$ and $\theta$. Specifically, step (1) maximizes $\tilde{Q}_s(\theta)$ over $s$ given $\theta = \theta_t$, and step (2) seeks to maximize $\tilde{Q}_s(\theta)$ over $\theta$ given $s$. This is similar to variational inference with $s$ being the variational parameter. %If we ignore the entropy term as discussed before, then it will be a joint maximization of $Q_s(\theta)$ over $s$ and $\theta$. 

The fixed point of the learning algorithm (\ref{eq:T}) solves the following estimating equation: 
\begin{align}
\frac{1}{n} \sum_{i=1}^{n} \E_{q_{s, \theta}(z_i|x_i)}\left[ \d \log p_{\theta}(x_i, z_i)\right] = 0. \label{eq:E}
\end{align}
If we approximate $\E_{q_{s, \theta_t}(z_i|x_i)}$ by Monte Carlo samples from $q_{s, \theta_t}(z_i|x_i)$, then the learning algorithm becomes Robbins-Monro algorithm for stochastic approximation \cite{robbins1951stochastic}. For fixed $s$, its convergence to the fixed point follows from regular conditions of Robbins-Monro. We expect that the optimized $s$ will also converge to a fixed value. 

It is worth stressing that $q_{s, \theta_t}(z_i|x_i)$ is the distribution under the short run MCMC. Thus fair samples can be obtained from $q_{s, \theta_t}(z_i|x_i)$ by running $K$ steps of short run MCMC. In contrast, the MLE estimating equation is $\frac{1}{n} \sum_{i=1}^{n} \E_{p_{\theta}(z_i|x_i)}\left[ \d \log p_{\theta}(x_i, z_i)\right] = 0$, where $p_{\theta}(z_i|x_i)$ is the posterior distribution. The MLE learning algorithm (\ref{eq:T0}) requires sampling from $p_{\theta_t}(z_i|x_i)$, which can be impractical, especially for multi-modal posterior distribution, where the mixing rate of MCMC can be very slow. %, whose convergence in finite time may require unrealistic or unverifiable theoretical assumptions, even for persistent chains. 

In our method, our estimate is defined by the solution to the estimating equation (\ref{eq:E}), which is a perturbation of the MLE estimating equation. We accept this bias, so that the learning algorithm can be justified as a Robbins-Monro algorithm, whose convergence can be easily established. %The theoretical underpinning of our learning method is thus much cleaner than the MLE learning algorithm. 
Thus both the target and the convergence of our learning algorithm are theoretically sound. 
 
The bias of the learned $\theta$ based on short run MCMC relative to the MLE depends on the gap between $q_{s, \theta}(z|x)$ and $p_\theta(z|x)$. We suspect that this bias may actually be beneficial in the following sense. The gradient ascent of $Q_s(\theta)$ seeks to increase $L(\theta)$ while decreasing  $\frac{1}{n}\sum_{i=1}^{n}  {\rm KL}(q_{s, \theta_t}(z_i|x_i) \| p_{\theta}(z_i|x_i))$. The latter tends to bias the learned model so that its posterior distribution $p_{\theta}(z_i|x_i)$ is close to the short run MCMC $q_{s, \theta_t}(z_i|x_i)$, i.e., our learning method may bias the model to make inference by short run MCMC accurate.

\section{Experiments}

In this section, we will demonstrate  (1) realistic synthesis, (2) faithful reconstructions of observed images, (3) inpainting of occluded images, (4) learning of hierarchical representations, (5) variational grid search and gradient descent on the step size, and, (6) ablation on latent layers and Langevin steps. The baselines are trained with ladder variational autoencoder~\cite{sonderby2016ladder} for multi-layer latent variable models. We refer to the Appendix and the reference implementation\footnote{\url{https://enijkamp.github.io/project_short_run_inference/}} for details.

% We emphasize the simplicity of the short run inference algorithm.

%All the training image datasets are resized and scaled to $[-1,1]$ with no further pre-processing. We train the models with $T=\num{3e5}$ parameter updates optimized by Adam \cite{kingma2015adam}. The learning rate $\eta$ decays step-wise (\num{1e-4}, \num{5e-5}, \num{1e-5}) for each \num{1e5} iterations. If not stated otherwise, we use $K=25$ short run inference steps and $\sigma$ is gradually annealed to $.15$.

\subsection{Synthesis}

\begin{figure}
	\begin{subfigure}{.5\textwidth}
		\centering
		\includegraphics[width=0.6\linewidth]{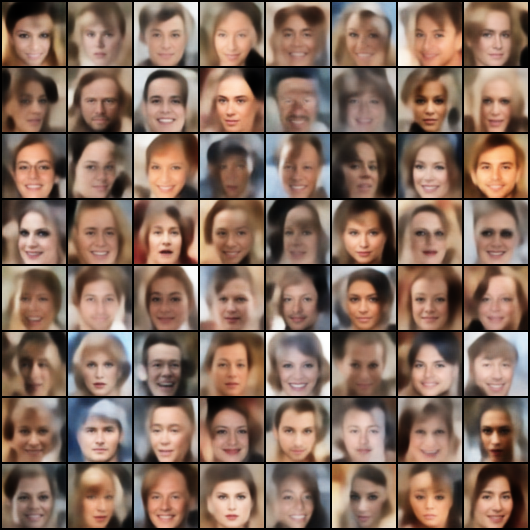}
		\caption{Ladder-VAE with $L=5$.}
		\label{fig:celeba64_vae}
	\end{subfigure}
	\begin{subfigure}{.5\textwidth}
		\centering
		\includegraphics[width=0.6\linewidth]{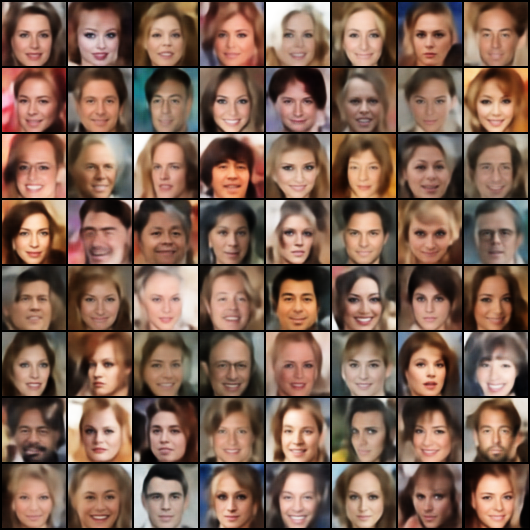}
		\caption{Short run inference with $K=25$.}
		\label{fig:celeba64_sri}
	\end{subfigure}
	\caption{Generated samples for models with $L=5$ layers on CelebA ($64\times64\times3$).}
	\label{fig:celeba64}
\end{figure}

\begin{figure}
	\centering
	\begin{subfigure}{.27\textwidth}
		\centering
		\includegraphics[width=.83\linewidth]{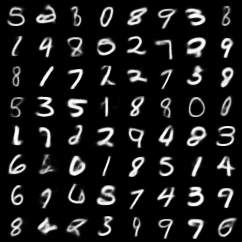}
		\caption{MNIST~($28 \times28$).}
	\end{subfigure}
	\begin{subfigure}{.27\textwidth}
		\centering
		\includegraphics[width=.83\linewidth]{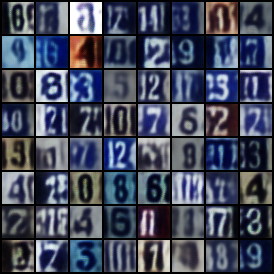}
		\caption{SVHN~($32\times32\times3$).}
	\end{subfigure}
	\begin{subfigure}{.27\textwidth}
		\centering
		\includegraphics[width=.83\linewidth]{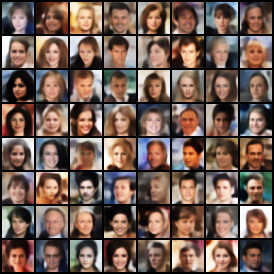}
		\caption{CelebA~($32\times32\times3$).}
	\end{subfigure}
	\caption{Generated samples for $K=25$ inference steps with $L=5$ layers.}
	\label{fig:synthesis}
\end{figure}

We evaluate the learned generator $g_\theta(z)$ by examining the fidelity of generated examples quantitatively on various datasets. Figure~\ref{fig:celeba64} depicts generated samples by our method and Ladder-VAE of size $64\times64$ pixels on the CelebA dataset. Figure~\ref{fig:synthesis} depicts generated samples of size $32\times32$ pixels for various datasets with $K=25$ short run MCMC inference steps. Table~\ref{tab:vae_abp} compares the Fr\'{e}chet Inception Distance~(FID)~\cite{heusel2017gans} with Inception v3 classifier~\cite{szegedy2016rethinking} on $40,000$ generated examples for the comparable multi-layer latent variable models Ladder-VAE~\cite{sonderby2016ladder} and Glow~\cite{kingma2018glow} for which levels may be comparable with layers of latent variables. Even though our method is specifically crafted for multi-layer latent-variable models, Table~\ref{tab:1_layer} compares short run MCMC on training single-layer latent-variable models with ABP~\cite{han2017abp}, GLO~\cite{boja2018glo}, VAE~\cite{kingma2013auto}, and VAE with MADE~\cite{germain2015made}. %For comparability, we trained the models with identical architecture for $g_\theta(z)$ and identical pre-processing of the datasets.
Despite its simplicity, short run MCMC is competitive with elaborate means of inference in VAE models and flow-based models, such as Glow~\cite{kingma2018glow}.

\begin{table}
	\footnotesize
	\begin{center}
		\begin{tabular}{cC{.3cm}C{1.2cm}C{1.2cm}C{.2cm}C{1.2cm}C{1.2cm}C{.2cm}C{1.2cm}C{1.2cm}} \toprule
			&& \multicolumn{2}{c}{MNIST} && \multicolumn{2}{c}{SVHN} && \multicolumn{2}{c}{CelebA}\\
			{Models} && {MSE} & {FID} && {MSE} & {FID} && {MSE} & {FID}\\ \midrule
			\text{Glow, $L=3$} && - & - && - & 65.27 && - & 39.84\\
			\midrule
			\text{Ladder-VAE, $L=1$} && 0.020 & - && 0.019 & 46.78 && 0.031 & 69.90\\
			\text{Ladder-VAE, $L=3$} && 0.018 & - && 0.015 & 41.72 && 0.029 & 58.33\\
			\text{Ladder-VAE, $L=5$} && 0.018 & - && 0.014 & 39.26 && 0.028 & 53.40\\
			\midrule
			\text{Ours, $L=1$} && 0.019 & - && 0.018 & 44.86 && 0.019 & 45.74\\
			\text{Ours, $L=3$} && 0.017 & - && 0.015 & 39.02 && 0.018 & 41.15\\
			\text{Ours, $L=5$} && 0.015 & - && 0.011 & 35.23 && 0.011 & 36.84\\
			\bottomrule
		\end{tabular}
	\end{center}
	\caption{Comparison of generators $g_\theta(z)$ with latent layers $L$ learned by Ladder-VAE and short run inference with respect to MSE of reconstructions and FID of generated samples for MNIST, SVHN, and CelebA $(32\times32\times3)$.}
	\label{tab:vae_abp}
% celeba
% L=1:
% short_run_abp/75_train_abp_mnist_32_hierarchical_ladder_one_celeba_1_one_deterministic.py
% L=5:
% 123_01_layers_five_celeba_3_bigger_with_noise_clamp_sigma_grid_more_steps_fix_sigma_lower_beta_lower_lr_redo118.py
% glow:
% 004_celeba_adamax_baseline.py
% 005_svhn_adamax_baseline.py
\end{table}

\begin{table}
	\footnotesize
	\begin{center}
		\begin{tabular}{C{3cm}C{1.7cm}C{1.7cm}C{1.7cm}C{1.7cm}C{1.7cm}} \toprule
			& ABP~\cite{han2017abp} & GLO~\cite{boja2018glo} & VAE~\cite{kingma2013auto} & VAE+IAF~\cite{germain2015made} & Ours\\ \midrule
			\text{SVHN} & 49.71 & 65.52 & 46.78 & 50.41 & 44.86\\
			\text{CelebA} & 51.50 & 50.70 & 69.90 & 53.78 & 45.74\\
			\bottomrule
		\end{tabular}
	\end{center}
	\caption{Comparison of generators $g_\theta(z)$ with latent layers $L=1$ with respect to FID of generated samples for SVHN and CelebA $(32\times32\times3)$.}
	\label{tab:1_layer}
\end{table}

\subsection{Reconstruction}

\begin{figure}[t]
	\begin{center}
		\includegraphics[width=.5\linewidth]{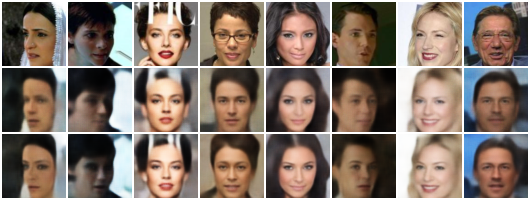}
	\end{center}
	\caption{Comparison of reconstructions between Ladder-VAE samples and our method on CelebA ($64\times64\times3$) with $L=5$. \textit{Top}: original test images. \textit{Middle}: reconstructions from VAE. \textit{Bottom}: reconstructions by short run inference.}
	\label{fig:reconstructions}
\end{figure}

We evaluate the accuracy of the learned short run MCMC inference dynamics $q_{s, \theta_t}(z|x_i)$ by reconstructing test images. In contrast to traditional MCMC posterior sampling with persistent chains, short run inference with small $K$ allows not only for efficient learning on training examples, but also the same dynamics can be recruited for testing examples. Figure~\ref{fig:reconstructions} compares the reconstructions of learned generators with $L=5$ layers by Ladder-VAE and short run inference on CelebA $(64\times64\times3)$. The fidelity of reconstructions by short run MCMC inference appears qualitatively improved over VAE, which is quantitatively confirmed by a consistently lower MSE in Table~\ref{tab:vae_abp}.

\subsection{Inpainting}

\begin{figure}[t]
	\setlength{\tabcolsep}{0.4pt} % Default value: 6pt
	\renewcommand{\arraystretch}{0.25} % Default value: 1
	\begin{center}
		\resizebox{.5\linewidth}{!}{
			\begin{tabular}{cccccccc}
				\includegraphics[width=0.11\linewidth]{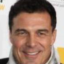} & 	\includegraphics[width=0.11\linewidth]{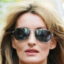} &
				\includegraphics[width=0.11\linewidth]{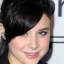} &
				\includegraphics[width=0.11\linewidth]{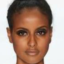} &
				\includegraphics[width=0.11\linewidth]{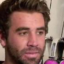} &
				\includegraphics[width=0.11\linewidth]{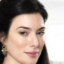} &
				\includegraphics[width=0.11\linewidth]{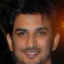} &
				\includegraphics[width=0.11\linewidth]{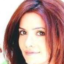} \\
				\includegraphics[width=0.11\linewidth]{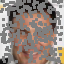} & \includegraphics[width=0.11\linewidth]{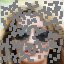} & 
				\includegraphics[width=0.11\linewidth]{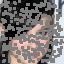} &
				\includegraphics[width=0.11\linewidth]{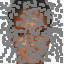} &
				\includegraphics[width=0.11\linewidth]{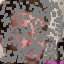} &
				\includegraphics[width=0.11\linewidth]{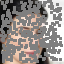} &
				\includegraphics[width=0.11\linewidth]{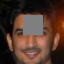} &
				\includegraphics[width=0.11\linewidth]{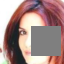} \\
				\includegraphics[width=0.11\linewidth]{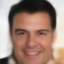} & \includegraphics[width=0.11\linewidth]{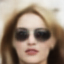} &
				\includegraphics[width=0.11\linewidth]{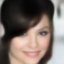} &
				\includegraphics[width=0.11\linewidth]{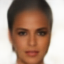} &
				\includegraphics[width=0.11\linewidth]{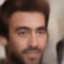} &
				\includegraphics[width=0.11\linewidth]{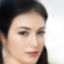} &
				\includegraphics[width=0.11\linewidth]{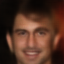} &
				\includegraphics[width=0.11\linewidth]{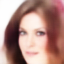} \\
			\end{tabular}
		}
	\end{center}
	\caption{Inpainting on CelebA ($64\times64\times3$) with $L=5$ for varying occlusion masks. \textit{Top}: original test images. \textit{Middle}: occluded images. \textit{Bottom}: inpainted test images by short run MCMC inference.}
	\label{fig:inpainting}
\end{figure}

Our method can ``inpaint'' occluded image regions. To recover the occluded pixels, the only required modification of (\ref{eq:S}) involves the computation of $\|x - g_\theta(z)\|^2/\sigma^2$. For a fully observed image, the term is computed by the summation over all pixels. For partially observed images, we only compute the summation over the observed pixels. Figure~\ref{fig:inpainting} depicts test images taken from the CelebA dataset for which a mask randomly occludes pixels in various patterns. 

% 93_train_abp_64_celeba_5_layers_32base_deconv_small_fc_cropped_grid_steps20_stepsize01_cont1_inpaiting.py
\setlength{\tabcolsep}{6pt} % Default value: 6pt
\renewcommand{\arraystretch}{1} % Default value: 1

\subsection{Hierarchical representation}

\begin{figure}[t]
	\centering
	\begin{subfigure}{.26\textwidth}
		\centering
		\includegraphics[width=0.9\linewidth]{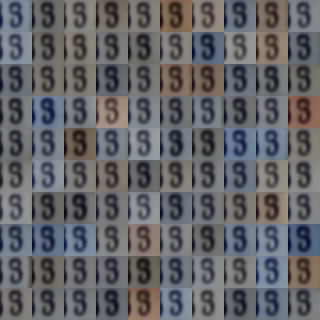}
		\caption{Bottom layer $z_1$.}
	\end{subfigure}
	\begin{subfigure}{.26\textwidth}
		\centering
		\includegraphics[width=0.9\linewidth]{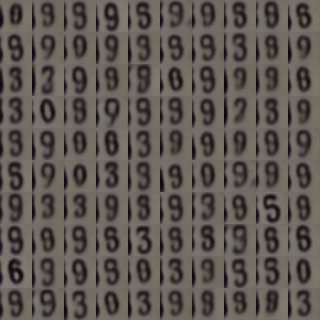}
		\caption{Middle layer $z_2$.}
	\end{subfigure}
	\begin{subfigure}{.26\textwidth}
		\centering
		\includegraphics[width=0.9\linewidth]{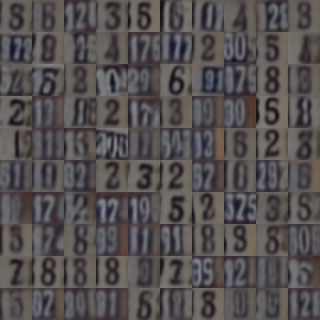}
		\caption{Top layer $z_3$.}
	\end{subfigure}
	\caption{Generated samples from a three-layer generator where each sub-figure corresponds to samples drawn when fixing the latent variables $z$ of all layers except for one. (a) The bottom layer represents background color. (b) The second layer represents digit identity. (c) The top layer represents general structure.}
	\label{fig:disentanglement}
\end{figure}

Multi-layer latent variable models not only demonstrate improved expressiveness over single-layer ones but also allow for learning the hierarchical structure. \cite{ermon17hierarch} argues that an alternative parameterization of the multi-layer generator promotes disentangled hierarchical features. We train a three-layer model with this parameterization using short run inference on SVHN. As shown in Figure~\ref{fig:disentanglement}, the three-layer latent variables capture disentangled representations, which are background color, digit identity, general structure from bottom to top layer.   

\subsection{Variational optimization of step size}

The step size $s$ in (\ref{eq:S}) may be optimized such that $q_s(z)$ best approximates the posterior $p_\theta(z|x)$. That is, we can optimize the step size $s$ by minimizing ${\rm KL}(q_s(z) \| p_\theta(z|x))$ via a grid search or gradient descent. As outlined in Section~\ref{sec:ss}, we require $d F(z_0, \epsilon)/d z_0$. In reverse-mode auto-differentiation, we construct the Jacobian one row at a time by evaluating vector-Jacobian products. %For large models, the Jacobian may be computed on CPU with memory of large capacity.
Then, we evaluate the eigenvalues of $d F(z_0, \epsilon)/d z_0$. As both steps are computed in a differentiable manner, we may compute the gradient with respect to $s$.

\begin{figure}[H]
	\centering
	\begin{subfigure}{.3\textwidth}
		\centering
		\includegraphics[width=1.0\linewidth]{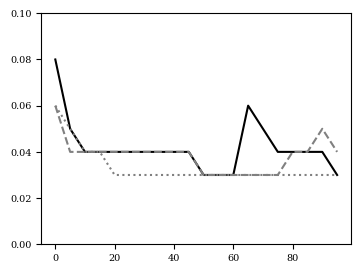}
		\caption{Grid-search.}
	\end{subfigure}
	\begin{subfigure}{.3\textwidth}
		\centering
		\includegraphics[width=1.0\linewidth]{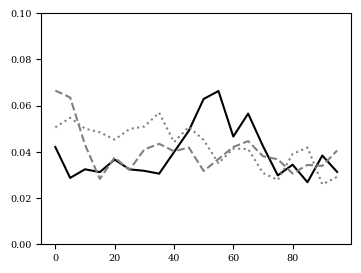}
		\caption{Gradient-descent.}
	\end{subfigure}
		\begin{subfigure}{.3\textwidth}
		\centering
		\includegraphics[width=1.0\linewidth]{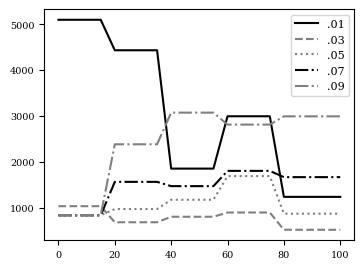}
		\caption{Gradient-descent.}
	\end{subfigure}
	\caption{(a) and (b) step size $s$ over epochs $T$ for three individual runs with varying random seed. (c) $\E_{q_s(z)}[\log q_s(z) -\log p_\theta(x, z)]$ for step sizes $s$ over epochs $T$.}
	\label{fig:ss1}	
\end{figure}

Figure~\ref{fig:ss1}a and \ref{fig:ss1}b depict the optimal step size $s$ over learning iterations $t$ determined by grid-search with $s \in \{0.01, 0.02, \ldots, 0.15\}$ and gradient descent on (\ref{eq:M}). For both grid-search and gradient descent the step size settles near $0.05$ after a few learning iterations. Figure~\ref{fig:ss1}c details the optimization objective of $s$, $\E_{q_s(z)}[\log q_s(z) -\log p_\theta(x, z)]$, with respect to individual step sizes $s$.

\subsection{Influence of number of layers and steps}

Tables~\ref{tab:L} and \ref{tab:K} show the influence of the latent layers $L$ for the generator network $g_\theta(z)$ and the number of steps $K$ in the inference dynamics (\ref{eq:S}), respectively. Increasing $L$ improves the quality of synthesis and reconstruction. Increasing $K$ up to 25 steps results in significant improvements, while $K>25$ appears to affect the scores only marginally.

\begin{table*}
	\footnotesize
	\centering
	\begin{subtable}{.41\textwidth}
		\centering
		\begin{tabular}{cccc}
			\specialrule{.1em}{.05em}{.05em}
			& \multicolumn{3}{c}{$L$}\\
			& $1$ & $3$ & $5$\\
			\hline
			FID & $61.03$ & $52.19$ & $47.95$ \\
			MSE & $0.020$ & $0.018$ & $0.015$ \\
			\specialrule{.1em}{.05em}{.05em} 
		\end{tabular}
		\caption{Varying $L$ with $K=25$.}
		\label{tab:L}
	\end{subtable}
% xsede:/home/enijkamp/pylon/short_run_abp/output/101_train_abp_64_celeba_1_layer_32base_deconv_small_fc_cropped_grid_steps20_stepsize01_cont1/2019-11-14-06-08-05/0
	\begin{subtable}{.55\textwidth}
		\centering
		\begin{tabular}{cccccc}
			\specialrule{.1em}{.05em}{.05em}
			& \multicolumn{5}{c}{$K$}\\
			& $5$ & $10$ & $25$ & $50$ & $400$\\
			\hline
			FID & $82.79$ & $67.38$ & $36.84$ & $35.39$ & $35.16$ \\
			MSE & $0.045$ & $0.037$ & $0.011$ & $0.010$ & $0.010$ \\
			\specialrule{.1em}{.05em}{.05em} 
		\end{tabular}
		\caption{Varying $K$ with $L=5$.}
		\label{tab:K}
	\end{subtable}
% enijkamp@single--p100-2:~/short_run_abp/output/100_ss5_5layers_celeba32_cont1/2019-11-15-02-32-34/0
	\caption{Influence of number of layers $L$ and number of short run inference steps $K$ on (a) CelebA ($64\times64\times3$) and (b) CelebA ($32\times32\times3$).}
\end{table*}

\section{Conclusion}

This paper proposes to use short run MCMC to infer latent variables in deep generative models, where the tuning parameters such as step size of the short run MCMC are optimized by a variational criterion. It thus combines the strengths of both MCMC and variational inference. Unlike variational auto-encoder, there is no need to design an extra inference model, which is usually a challenging task for models with multiple layers of latent variables. 

The short run MCMC is easily affordable on the current computing platforms and can be easily scaled up to big training data. It will enable the researchers to develop more sophisticated latent variable models, such as biologically plausible models where each node is a latent variable and the short run MCMC can be compared to attractor dynamics in neuroscience. 

This paper lays the foundation for short run MCMC for approximate inference in complex generative models, where the short run MCMC  is optimized in a principled way. In our further work, we shall explore more efficient approximate methods for optimizing or learning more general forms of short run inference dynamics. 

\subsubsection*{Acknowledgments}
The work is supported by NSF DMS-2015577, DARPA XAI N66001-17-2-4029,  ARO W911NF1810296, ONR MURI N00014-16-1-2007, and XSEDE grant ASC170063. We thank NVIDIA for the donation of Titan V GPUs. We thank Eric Fischer for the assistance with experiments.

\clearpage

\bibliographystyle{splncs04}
\bibliography{egbib}

\clearpage
\section{Appendix}

\subsection{Experiment details}
All the training image datasets are resized and scaled to $[-1,1]$ with no further pre-processing. We train the models with $T=\num{3e5}$ parameter updates optimized by Adam \cite{kingma2015adam}. The learning rate $\eta$ decays step-wise (\num{1e-4}, \num{5e-5}, \num{1e-5}) for each \num{1e5} iterations. If not stated otherwise, we use $K=25$ short run inference steps and $\sigma$ is gradually annealed to $0.15$.

\subsection{Model specification}

For the multi-layer generator model, we have $z=\left(z_l,l=1,\ldots,L\right)$ for which layer $L$ is the top layer, and layer $1$ is the bottom layer close to $x$. For simplicity, let $x=z_0$. Then, $p_\theta(z) = p_\theta(z_L)\prod_{l=0}^{L-1}p_\theta(z_l\mid z_{l+1})$. In our case, we have $z_L \sim \N(0, I)$, 
$[z_l|z_{l+1}] \sim \N(\mu_l(d_l(p_l(z_{l+1}))), \sigma_l^2(d_l(p_l(z_{l+1})))),  \; l = 0, ..., L-1$.
where $\mu_l()$ and $\sigma_l^2()$ are the mean vector and the diagonal variance-covariance matrix of $z_l$ respectively, and they are functions of $d_l(p_l(z_{l+1}))$ where $d_l$ are deterministic layers and $p_l$ are projection layer to preserve dimensionality. $d_l$ is defined as two subsequent $conv2d$ layers with $GeLU$ \cite{hendrycks2016gelu} activation functions and skip connection. $p_l$ is a linear layer with subsequent $transpose\_conv2d$. $\mu_l$ and $\sigma_l$ are a pair of $conv2d$ and $linear$ layers to project to dimensionality of $z_{l}$. Then, $z_l = \mu_l(d_l(p_l(z_{l+1}))) + \sigma_l(d_l(p_l(z_{l+1}))) \otimes \epsilon_l$ where $\epsilon_l \sim \N(0, I_{d_l})$. The final deterministic block $o_0$ is a $transpose\_conv2d$ layer projecting to the desired dimensionality of $x$. The range of $x$ is bounded by $tanh()$.

Table~\ref{tab:spec} illustrates a specification with $L=3$ latent layers, latent dimensions $d_3=32$, $d_2=64$, $d_1=128$ for $z_3$, $z_2$, $z_1$, respectively, and $n_f=64$ channels.

\begin{table}[h!]
	\begin{center}
		\begin{tabular}{ |c|c|l| } 
			\hline
			$l$ & operation & dimensions \\
			\hline
			3 & $z_3\sim N(0, I_{d_3})$ & $[n,d_3,1,1]$ \\ 
			\hline 
			2 & $z_{3,p} = p_2(z_3)$ & $[n,n_f,16,16]$ \\ 
			\hline
			2 & $z_{3,d} = d_2(z_{3,p})$ & $[n,n_f,16,16]$ \\ 
			\hline
			2 & $z_2 = \mu_2(z_{3,d}) + \sigma_2(z_{3,d}) \otimes \epsilon_2$ & $[n,d_2,1,1]$ \\ 
			\hline
			1 & $z_{2,p} = p_1(z_2)$ & $[n,n_f,16,16]$ \\ 
			\hline
			1 & $z_{2,d} = d_1(z_{2,p}) + z_{3,d}$ & $[n,n_f,16,16]$ \\ 
			\hline
			1 & $z_1 = \mu_1(z_{2,d}) + \sigma_1(z_{2,d}) \otimes \epsilon_1$ & $[n,d_1,1,1]$ \\ 
			\hline
			0 & $z_{1,p} = p_0(z_1)$ & $[n,n_f,16,16]$ \\ 
			\hline
			0 & $z_{1,d} = d_0(z_{1,p}) + z_{2,d}$ & $[n,n_f,16,16]$ \\ 
			\hline
			0 & $x = tanh(o_0(z_{1,d}))$ & $[n,3,32,32]$ \\
			\hline
		\end{tabular}
		\medskip
		\caption{Specification of multi-layer generator model with $L=3$ layers, latent dimensions $d_3=32$, $d_2=64$, $d_1=128$ for $z_3$, $z_2$, $z_1$, respectively, and $n_f=64$ channels.}
		\label{tab:spec}
	\end{center}
\end{table}

\subsection{Training of baselines}

For ladder variational autoencoder~\cite{sonderby2016ladder}, the generator model is defined in Table~\ref{tab:spec}. The training follows the one outlined in~\cite{sonderby2016ladder}. We train the model with $T=\num{3e5}$ parameter updates optimized by Adam \cite{kingma2015adam}. 

For GLO~\cite{boja2018glo} and ABP~\cite{han2017abp}, our model in Table~\ref{tab:spec} was reduced to a single-layer variational autoencoder.

For GLO, we used a re-implementation\footnote{\url{https://github.com/tneumann/minimal_glo}} in PyTorch. As outlined in \cite{boja2018glo}, after training the model, the inferred latent vectors, $z$, were used to fit a multivariate Gaussian distribution from which $z$ was drawn for sampling. The hyperparameters are as follows: $code\_dim=128$, $n\_pca=64 * 64 * 3 * 2$, $loss=l2$.

For ABP, $40$ steps of persistent Markov Chains were used. The hyper-parameters are as follows: $40$ MCMC steps, Langevin discretization step size of $0.3$, $\sigma=0.3$, Adam~\cite{kingma2015adam} optimizer.

For Glow~\cite{kingma2018glow}, the model was trained using the official code\footnote{\url{https://github.com/openai/glow}} with our datasets and the evaluation was performed with our implementation of the Fréchet Inception Distance~(FID)~\cite{heusel2017gans} with Inception v3 classifier~\cite{szegedy2016rethinking} on $40,000$ generated example. The hyperparameters are as follows: $dal=0$, $n\_batch\_train=64$, $optimizer=adamax$, $n\_levels=3$, $width=512$, $depth=16$, $n\_bits\_x=8$, $learntop=False$,  $flow\_coupling=0$.

\end{document}